\documentclass[conference]{IEEEtran}
\IEEEoverridecommandlockouts
\usepackage{cite}
\usepackage{amsmath,amssymb,amsfonts}
\usepackage{algorithmic}
\usepackage{graphicx}
\usepackage{textcomp}
\usepackage{xcolor}
\def\BibTeX{{\rm B\kern-.05em{\sc i\kern-.025em b}\kern-.08em
    T\kern-.1667em\lower.7ex\hbox{E}\kern-.125emX}}
\begin{document}
\title{Stability Analysis of ChatGPT-based Sentiment Analysis in AI Quality Assurance
}
%\author{\IEEEauthorblockN{Anonymous Authors}}
\author{
Tinghui Ouyang$^{1}$, 
%Hoang-Quoc Nguyen-Son$^{1}$,
AprilPyone MaungMaung$^{1}$, Koichi Konishi$^{2}$,\\
Yoshiki Seo$^{2}$, and Isao Echizen$^{1}$ \\
\small{\textit{$^{1}$National Institute of Informatics }} \\
\small{\textit{$^{2}$National Institute of Advanced Industrial Science and Technology, Japan}} \\
\small{E-mail: \{thouyang, pyone, iechizen\}@nii.ac.jp; \{k.konishi, y.seo\}@aist.go.jp}
}

\maketitle

\begin{abstract}
In the era of large AI models, the complex architecture and vast parameters present substantial challenges for effective AI quality management (AIQM), e.g. large language model (LLM). This paper focuses on investigating the quality assurance of a specific LLM-based AI product—a ChatGPT-based sentiment analysis system. The study delves into stability issues related to both the operation and robustness of the expansive AI model on which ChatGPT is based. Experimental analysis is conducted using benchmark datasets for sentiment analysis. The results reveal that the constructed ChatGPT-based sentiment analysis system exhibits uncertainty, which is attributed to various operational factors. It demonstrated that the system also exhibits stability issues in handling conventional small text attacks involving robustness.
\end{abstract}

\begin{IEEEkeywords}
AI quality management, large language model (LLM), ChatGPT-based sentiment analysis, stability analysis, 
\end{IEEEkeywords}

%\documentclass[10pt,a4paper]{article}
%\usepackage[utf8]{inputenc}
%\usepackage{amsmath}
%\usepackage{amsfonts}
%\usepackage{amssymb}
%\usepackage{graphicx}
%\begin{document}
%
%
%\title{
%}
%\author{Tinghui Ouyang}
%\maketitle
%
%\begin{abstract}
%Abstract:xxxx
%\end{abstract}
%keywords:

\section{Introduction}

Advances in machine learning, especially in deep learning (DL), have led to the application of artificial intelligence (AI) to a broad range of areas, e.g., autonomous driving, e-commerce, and robotics \cite{b1}-\cite{b3}. However, due to the black-box property of DL models and the lack of effective evaluation and testing techniques, it is difficult to identify the cause of malfunctions in AI-based products. The study of AI quality management (AIQM) \cite{b4} is aimed at providing useful techniques for AI-based product evaluation, testing, and improvement. It typically encompasses various aspects related to internal and external qualities, such as  data- and model-related qualities. Therefore, AIQM is critical for ensuring the effectiveness and reliability of AI-based products.

The increasing power of computer devices has made possible the development of complex AI architectures with a great number of parameters. Thus, many large models have been developed in the fields of computer vision \cite{b5} and natural language processing (NLP) \cite{b6}. The most well-known example is ChatGPT \cite{b7}, which was released by OpenAI in 2023. It can generate human-like text with context and coherence. Its ability to understand natural language, even the nuances of language, and to generate highly accurate responses is the result of pretraining on a massive amount of text data using an advanced neural architecture (i.e., a Generative-Pretrained--Transformer (GPT) \cite{b8}) with a great number of parameters. As an advanced cutting-edge AI product currently in the limelight, ChatGPT has attracted much attention due to its superiority in addressing various traditional NLP tasks as well as its generation ability. For example, ChatGPT can deal with various languages besides English, so it was used as a translator in \cite{b9}. By making use of its ability for semantic understanding, ChatGPT was also used as a grammatical checker \cite{b10}. Its most widely used ability is conversation ability, enabling it to be used as a chatbot for question answering, or as a tool to guide educational exams \cite{b11}, and to assist diagnosis in medical counseling \cite{b12}. Moreover, by making use of ChatGPT's generation ability, cheating in writing is a serious problem in education \cite {b13}. In \cite{b14}, ChatGPT was also used as a tool to augment text data for other research.

Along with the convenience and benefits of ChatGPT, it also poses significant challenges to our society, including not only the agnosticism and concerns to everyday people  but also the difficulties and complexities presented to researchers and developers. Especially, researchers in software engineering and quality management in particular need to understand how ChatGPT works, how to evaluate it, and how to maintain it.
Even though it has been accessible to users for only a short while, much research has been conducted on the quality management of ChatGPT due to its prominence and wide usage. For example, in terms of the correctness study, ChatGPT's performances have been evaluated on several traditional NLP tasks, like translation, grammar checking, text animation, and classification, and compared with that of conventional products. Moreover, Other studies have evaluated its performance on novel tasks like mathematical capabilities \cite{b15} and logical reasoning in coding \cite{b16}. For the robustness study, some adversarial text data in various NLP tasks were used to evaluate ChatGPT’s performance \cite {b17}-\cite {b18}. Some researchers also tried to attack LLM for robustness analysis, for example, using the gradient-based method for a prompt attack on Llama and other open-sourced LLMs \cite{b19}, but it is hard to perform such analysis on ChatGPT because of its black-box property. In \cite{b20}, an interesting idea was proposed to utilize LLM to generate attacks and then test its robustness on these attacks. Moreover, some other quality studies have also been performed. For example, the reliability of ChatGPT on text classification was studied in \cite{b21}. The limitations and weaknesses  of ChatGPT faced with some questions were also reported, like political questions \cite{b22}, failure cases \cite{b23}, and so on \cite{b24}-\cite{b25}.

However, due to ChatGPT's complex architecture and black-box properties, there are still many difficulties. Evaluating ChatGPT directly on the basis of these related studies is challenging \cite{b26}. Since ChatGPT can deal with many tasks based on the foundation model, it has no clear function requirements as addressed in traditional software quality management. A useful way is to concentrate on a specific NLP task for quality analysis, e.g., accuracy on sentiment analysis, grammar checking, and translation. The other possible way is to use some surrogate datasets to evaluate ChatGPT's performance on specific qualities, e.g., applying an existing adversarial dataset, AdvGLUE \cite{b27}, to test the robustness of ChatGPT. These
solutions can evaluate and find some weaknesses of ChatGPT, however its shortages are not analyzed directly as a software product.

Given these considerations, this paper proposes to study the stability quality of a ChatGPT-based software product, namely a ChatGPT-based sentiment analysis system. Sentiment analysis was selected because it is the simplest task in NLP and because it has a clear problem formulation and evaluation metrics for quality assurance \cite{b28}-\cite{b29}. Moreover, this study adopts an architecture oriented to a specific task and a specific quality. It focused on the causes of instability from the perspective of software engineering and on robustness against attacks. Experiments using benchmark sentiment analysis datasets demonstrated that ChatGPT is robust against attacks when used for sentiment analysis although weaknesses were revealed for synonym perturbations.

The rest of this paper is organized as follows. Section 2 gives an overview of using ChatGPT for sentiment analysis and describes the two aspects of stability in AIQM. Based on the division of stability studies, Section 3 and 4 analyze the stability issues of ChatGPT from different aspects, such as uncertainty causes of ChatGPT operation in Section 3 and the AI model's robustness issues in Section 4. Section 5 concludes the work of this paper.

\section{Overview}
\subsection{ChatGPT-based sentiment analysis system}
As described above, our major objective was to investigate  the AI quality assurance of a ChatGPT-based sentiment analysis system. It is known that the product's functions are clear in traditional software testing. However, ChatGPT is based on the foundation model which can address several practical tasks e.g., question answering, chatting, text generation, grammar checking, and translation. To evaluate the quality of ChatGPT as an AI product, we need to investigate multiple aspects. This is a huge time-consuming task. To simplify this AIQM task, we focused on evaluating ChatGPT's ability on a specific NLP task, namely sentiment analysis. In this paper, sentiment analysis is selected because it is a simple and typical NLP task. Moreover, it is a classification problem for which there are standard evaluation metrics. Therefore, the objective of this paper in AIQM is to evaluate the quality of the ChatGPT-based sentiment analysis system.

\begin{figure}
\centering
\includegraphics[scale=0.5]{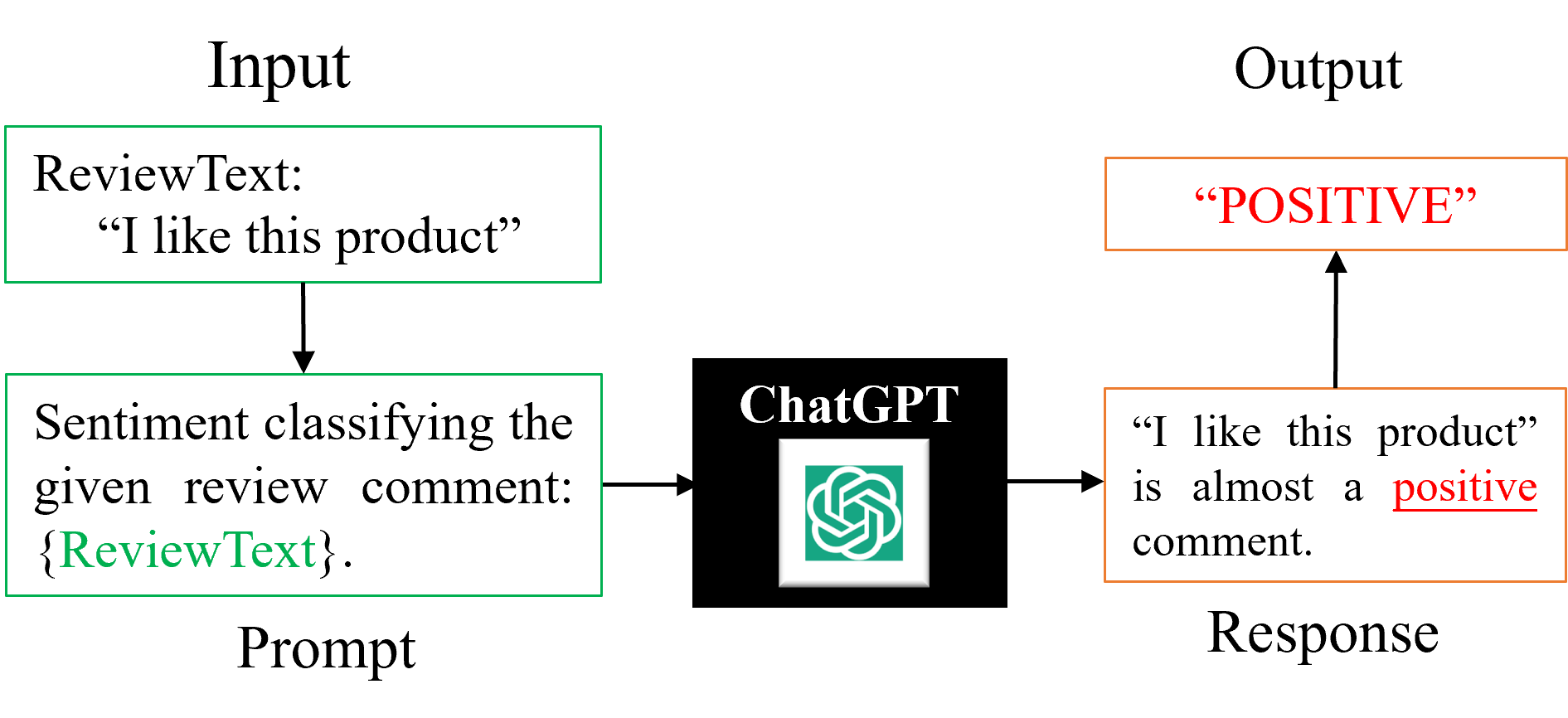}
\caption{Diagram of using ChatGPT for sentiment analysis}
\label{f1}
\end{figure}
The diagram in Fig. \ref{f1} illustrates how using ChatGPT for sentiment analysis. Compared with conventional testing based on input and output data, e.g., review comments and sentiment labels in sentiment analysis, it is found that some differences happen in the process of using ChatGPT in Fig. \ref{f1}. Since ChatGPT operates in conversation mode, extra editing on prompt and response is usually needed to make ChatGPT focus on a specific task and produce the desired output. An understanding of this difference, and with the help of the ChatGPT API provided by OpenAI, it becomes easy for users to develop a specific AI product based on ChatGPT. The following are details about the settings for developing a ChatGPT-based sentiment analysis system in our AIQM study.
\vspace{0.5cm}
\begin{itemize}
\item PromptSetting: Analyze the following product review and determine if the sentiment is POSITIVE, NEGATIVE or NEUTRAL: \{ReviewText\}
\item OutputControl: Return only a single word, such as POSITIVE, NEGATIVE or NEUTRAL.
\end{itemize}
\vspace{0.5cm}

\subsection{Stability of AI}
According to the Machine Learning Quality Management guideline \cite{b4}, among the diverse AI qualities, stability is a crucial quality that deserves special attention. In this context, stability refers to an AI product  performing consistently and reliably. It encompasses the ability of the product to operate seamlessly under varying conditions and to resist disruptions. This means that the product must be less prone to errors, crashes, or unexpected behaviors. It is thus crucial to achieve and maintain AI product stability to ensure a positive user experience and to minimize the effect of potential problems.

Moreover, with respect to AI-based software products, the stability study can be approached in two ways. One is to focus on system stability, specifically the intricacies of operational stability and the challenges posed by uncertainty. The second is to focus on model stability, mainly related to the AI model used in the software product. In other words, the second type of stability is about the robustness analysis of trained AI models.In the subsequent sections, we will explore these two stability studies in the context of ChatGPT in detail.

\section{Uncertainty analysis}
\subsection{Model architecture design}
\begin{figure}
\centering
\includegraphics[scale=0.3]{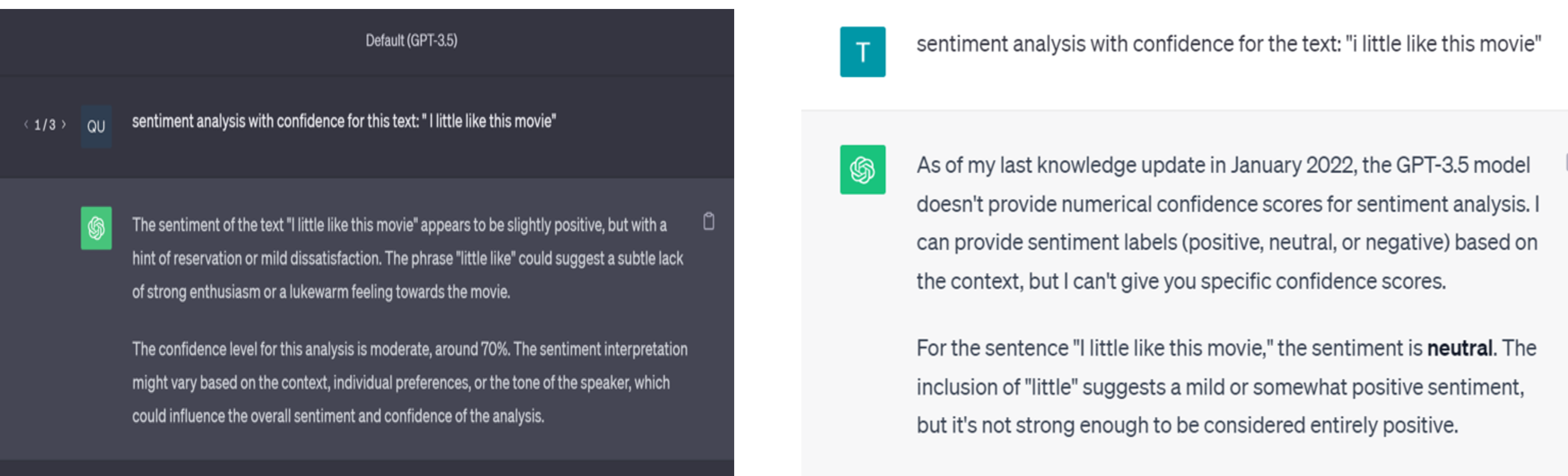}
\caption{ChatGPT’s responses on two devices at same time}
\label{f2}
\end{figure}

Currently, it is widely discussed that uncertainty issues exist in the running of ChatGPT. Both with the advanced GPT4 or the widely used GPT-3.5-turbo model, the responses are non-deterministic even for a temperature setting of 0.0. For example, repeated inputs of a question to ChatGPT generally produces different responses. One possible reason for this is that ChatGPT is continually being updated on the basis of data collected from global customers. However, as shown in Fig. \ref{f2}, the responses on two different devices for the same input at the same time differed. Although the responses could be affected by randomness in the text generation process, the major reason for the non-determinism of responses is the sparse MoE (mixture of expert models) architecture in ChatGPT \cite{b30}. This architecture organizes tokens into fixed-size groups due to capacity constraints and emphasizes balance within each group. When groups incorporate tokens from diverse sequences or inputs, there is often competition between expert buffers among these tokens, leading to a failure in enforcing per-sequence determinism.

\subsection{Difference on using ChatGPT and ChatGPT API} 
Developers using ChatGPT have reported differences between using ChatGPT on the web  and the ChatGPT API. A general finding is that the web version  performs better than the API even when using the same model. Several reasons for this have been suggested \cite{b31}–\cite{b32}. One is that the web version is continually being updated on the basis of the huge amounts of data being received from global customers, whereas the API version is fixed for a certain period of time. Another is related to the system prompt. ChatGPT has a default system prompt, namely " You are an LLM-based AI system created by OpenAI ..." or what model it is using, or something else . However, when using the API for testing, developers must construct their own system prompt to specify how ChatGPT should behave. For example, when using the API to develop a sentiment analysis system, the system prompt can be set as 

\textit{"You are an AI language model trained to analyze and detect the sentiment of product reviews"}

Moreover, whereas ChatGPT is a conversation-based system which can collect historical data for AI to produce the most satisfactory results, the API is a simple question-answer chat robot. A conversation management system must be created to enable history-based conversation. Another possible reason that the web version  performs better than the API even when using the same model is a difference in the setting of the undisclosed “temperature” parameter in ChatGPT. This parameter controls the degree of creativity or unpredictability in response generation. Generally, the higher the value, the more uncertain the result. To obtain a stable result, a temperature setting of 0.0  should be used.

%ChatGPT has a conversation management system that will pass recent relevant chat back to the AI so it can see what you were recently discussing.
%The API requires you to write the conversation management system yourself to allow the AI to see prior conversation besides just the latest input if you want this chat behavior.

%ChatGPT has an undisclosed “temperature” parameter, which controls how creative or unpredictable the generation of words is;
%In the API, you will likely want temperature 0.5 for most uses, high enough that it doesn’t always give the same answer, but low enough it isn’t completely improvisational.
%With identical model, identical system prompt, and identical user input, you will get the same output from the API. Your conversation history can be even better than ChatGPT after many turns if you want to pay the extra amount for submitting maximum chat possible along with every new question.

%moreover, eventhough the same prompt at different device, it may has different results
%\begin{figure}
%\centering
%\includegraphics[scale=0.5]{fig14.png}
%\caption{results of different time}
%\end{figure}
\subsection{Variance due to timing}
\begin{figure}
\centering
\includegraphics[scale=0.35]{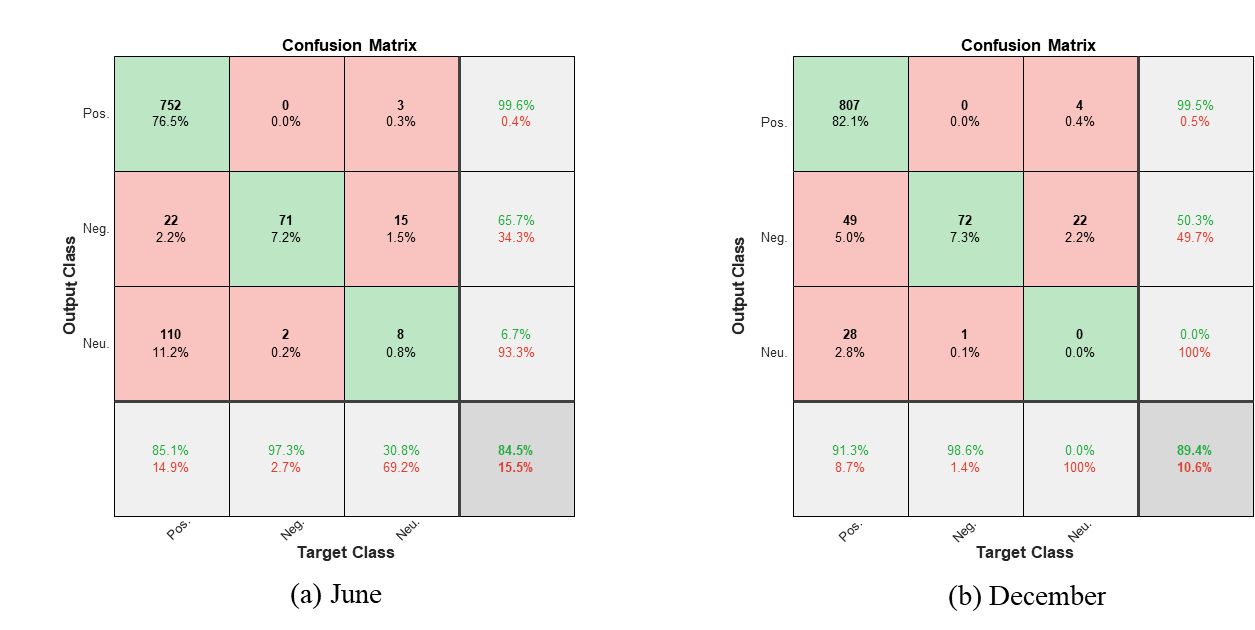}
\caption{ChatGPT for sentiment analysis at different time}
\label{f4}
\end{figure}

As discussed above, ChatGPT continuously updates itself on the basis of newly collected data and design changes. For example, OpenAI has made both GPT-3 and GPT-4 models available for use in ChatGPT and reports that GPT-4 performs better than GPT-3 on most tasks. Even for two identical ChatGPT systems processing the same input at the same time, there is uncertainty about the output. Since the process for updating the models is not transparent, the effect of each update on model behavior is unclear. These uncertainties pose challenges in studying the stability of ChatGPT. For example, if there is a sudden change in a model's response to a prompt (e.g., its accuracy or formatting), downstream processing may be disrupted. Furthermore, it is challenging to reproduce a model’s results even with the same settings. This issue has been discussed on the basis of comparative experiments on several NLP downstream tasks \cite{b33}.
In this paper, using the same set-up, performances of using ChatGPT for sentiment analysis at different time slots are also experimented with, as shown in Fig. \ref{f4}.

A comparison of the results using the ChatGPT API for sentiment analysis in June and December 2023 is shown in Fig. \ref{f4}. The dataset was constructed using Amazon.com review data, and the settings were the same. The confusion matrices show that the June version was better able to comprehend the prompts and thus better able to distinguish all positive, negative, and neutral reviews. Updating appears to have affected the performance of the December version: it focused more on polar classification (positive or negative), reflecting weakness in recognizing neutral review comments, as shown in Fig. \ref{f4}. This change in ChatGPT's ability to perform sentiment analysis means that it is essential to determine what happens when the model version of ChatGPT is updated.

\subsection{Prompt engineering} 
\begin{figure}
\centering
\includegraphics[scale=0.35]{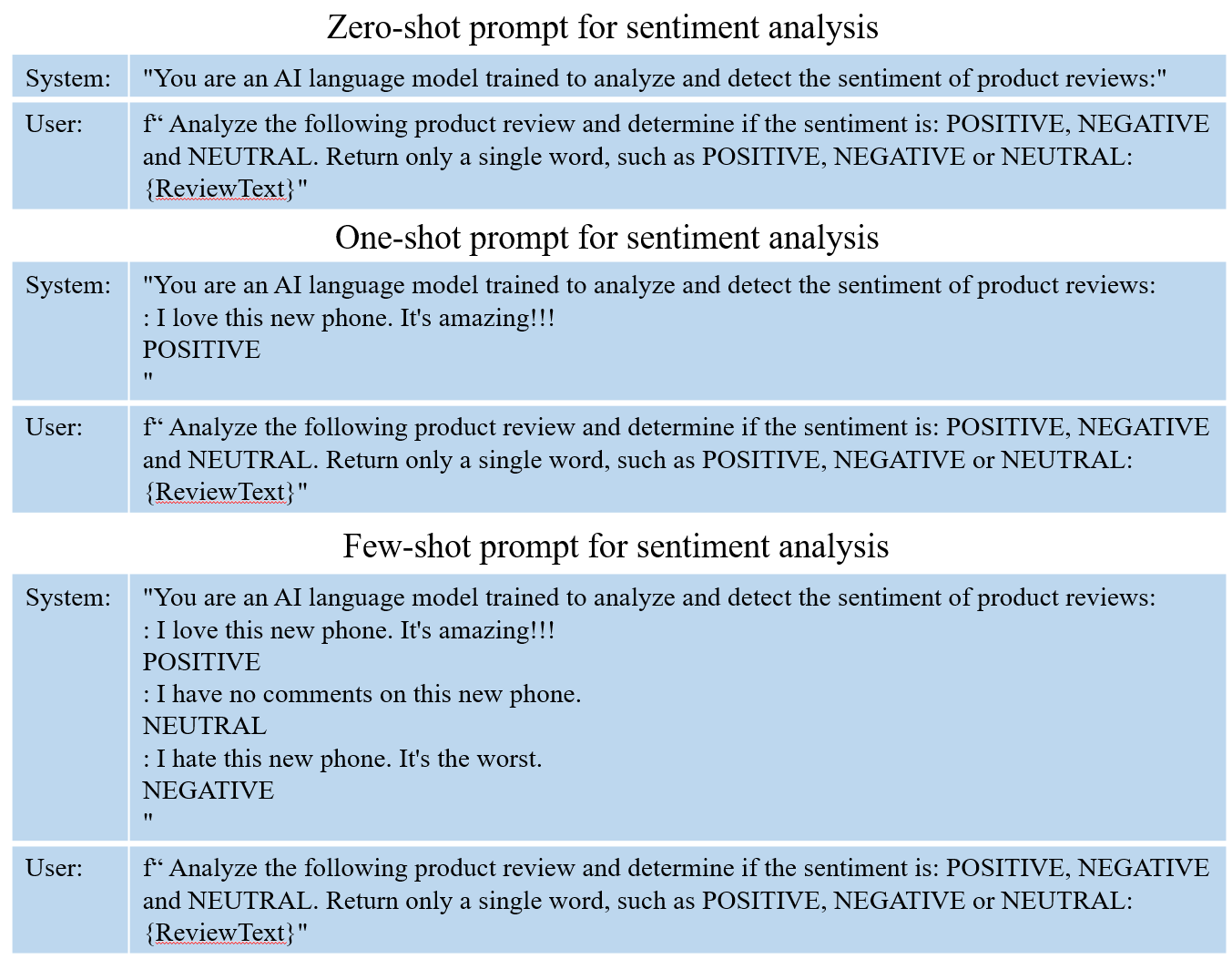}
\caption{Designs of zero-, one-, and few-shot prompts for sentiment analysis }
\label{f5}
\end{figure}

In the operation of ChatGPT, another issue that affects the stability of ChatGPT operation is the prompts. Prompt engineering is an essential topic in the study of LLMs. Prompts should usually be carefully designed to optimize the output. To investigate how the prompts affect the stability of ChatGPT, this paper designed several prompt settings. By making use of the in-context learning ability of ChatGPT, the prompt engineering here is set based on the criterion of zero-shot, one-shot, and few-shot, as shown in Fig. \ref{f5}.

\begin{figure*}
\centering
\includegraphics[width=0.8\textwidth]{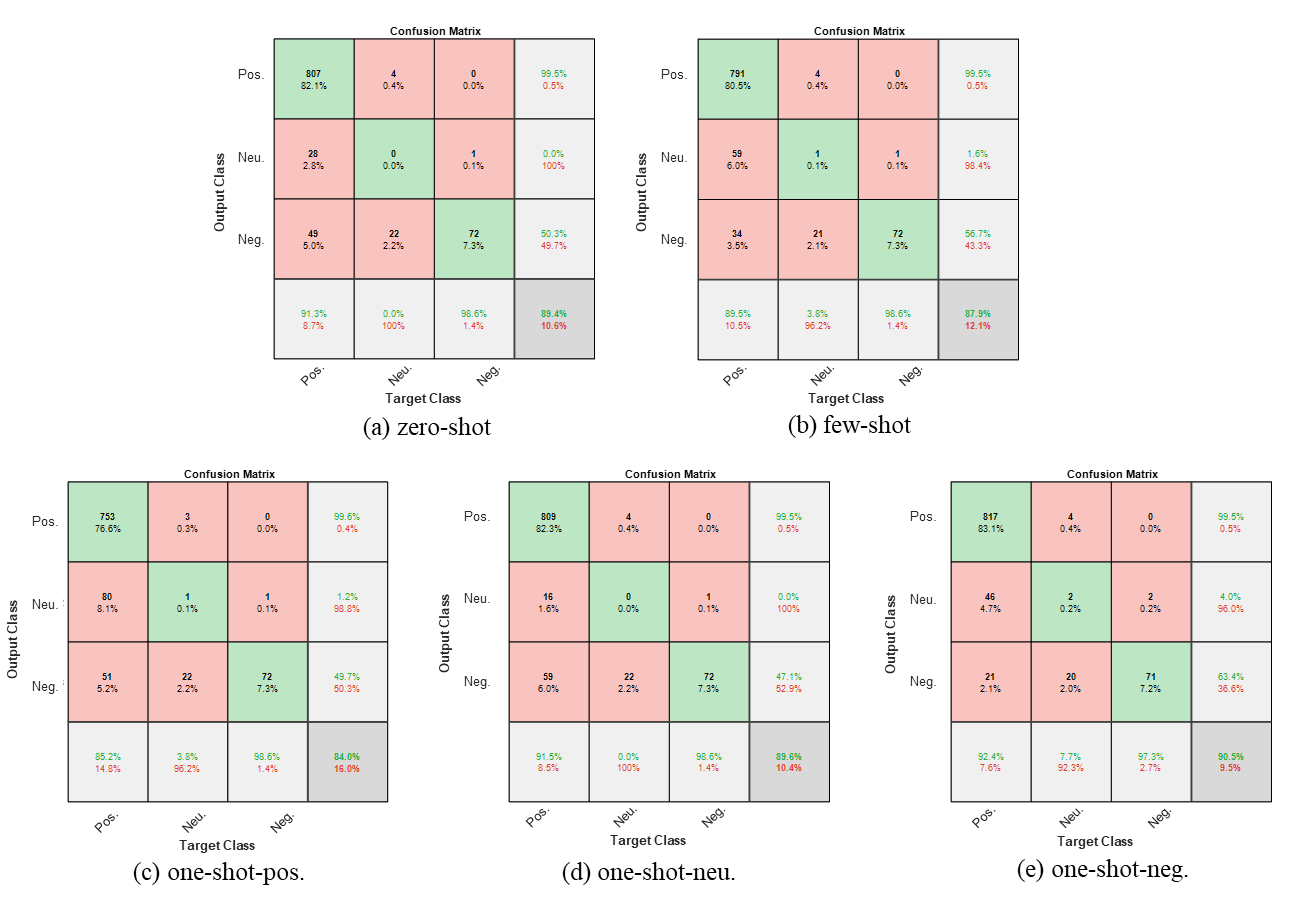}
\caption{Sentiment analysis results for different prompt settings}
\label{f6}
\end{figure*}

In Fig. \ref{f5}, three example prompts are given for using ChatGPT in sentiment analysis. It is seen that with a zero-shot prompt, no example information is provided, and with a few-shot prompt, several examples are provided for guidance. A one-shot prompt is a special case of a few-shot prompt where only one example is provided for guidance. Therefore, based on this idea, three sub-designs providing positive, neutral, and negative examples, respectively, are also considered in this paper. Also, based on Amazon.com review data for sentiment analysis, the evaluation results are presented via the confusion matrix, as shown in Fig. \ref{f6}.

As shown by the confusion matrices in Fig. \ref{f6}, different prompts resulted in different sentiment classification accuracies for the same dataset, indicating that prompt engineering indeed affects the stability of ChatGPT performance. The accuracy with the zero-shot prompt was slightly better than that with the few-shot prompt. The main difference was that the few-shot prompt reduced prediction accuracy for positive reviews. Looking at the matrices for the three one-shot prompts with positive/neutral/negative examples, we see that using the ones with neutral and negative examples seems to have improved accuracy, indicating that the performance of ChatGPT for sentiment analysis can be improved by careful engineering of the prompts.

\section{robustness testing }

According to the above description, the second type of stability analysis is model stability, specifically the robustness of the trained AI model. Given that the focus of our study was the use of ChatGPT for sentiment analysis, we investigated the robustness of the AI model on which it is based.
 
The IEEE standard notation in software engineering terminology \cite{b34} indicates that robustness is the degree to which a system or component can function correctly in the presence of invalid inputs and/or stressful environmental conditions. In general, invalid inputs are data generated by perturbation or mutation, and a stressful environmental condition would be, for example, an attack or threat. Therefore, investigation of an AI model’s robustness involves an in-depth analysis of its resilience to perturbations. With respect to the ChatGPT-based model, we comprehensively evaluated  model robustness by taking into account potential vulnerabilities arising from different types of perturbation. Such an evaluation contributes to a deeper understanding of the interplay between AIQM and provides insights into ways to enhance the robustness and reliability of AI systems.

%emoji and promp-
%\begin{table}
%\centering
%\caption{comparison of different perturbations}
%\hline
%Naturalness& Appearance & Meaning & pronunciation & Embedding\\
%Typo& \surd &
%
%\end{table}

\subsection{Data preparation}
In this paper, two commonly used datasets for sentiment analysis, namely the Amazon.com review dataset \cite{b35} and the Stanford Sentiment Treebank (SST) dataset \cite{b36}, are used for the evaluation. 

\begin{itemize}
\item Amazon.com review dataset: This dataset is a collection of a large number of product reviews from Amazon.com. The raw data contains 82.83 million unique reviews and includes product and user information, a rating score (1–5 stars), and a plain text review. For sentiment analysis, researchers usually take a review score of 1 or 2 as negative, 4 or 5 as positive, and 3 as neutral. We did likewise.
\item SST dataset: This dataset consists of 11,855 individual sentences extracted from movie reviews by Pang and Lee \cite{b36}. Applying the Stanford parser to this dataset enables a comprehensive examination of the compositional effects of sentiment in language. In this paper, we used an extension of this dataset with fine-grained labels (very positive, positive, neutral, negative, very negative) and roughly categorized the review sentiments as positive, neutral, or negative. 
\end{itemize}

\begin{figure}
\centering
\includegraphics[scale=0.3]{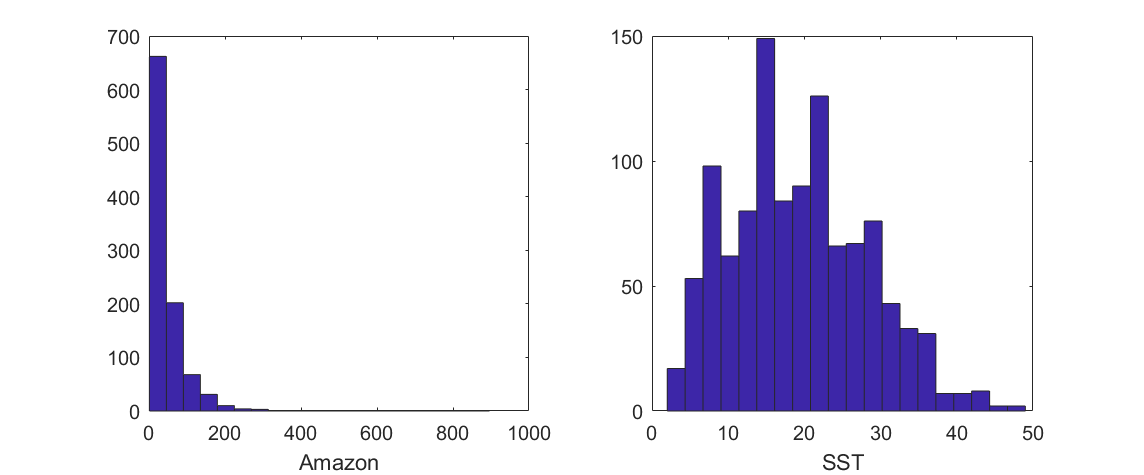}
\caption{Length of reviews in Amazon.com review and SST datasets}
\label{f7}
\end{figure}

In this paper, only a small amount of samples selected from the original benchmark datasets are taken as the testing sets in evaluation. Table \ref{tb1} presents the information of selected testing sets. 
\begin{table}
\centering
\caption{Dataset information}
\begin{tabular}{cccc}
\hline
Dataset & No. of samples & Distribution (Pos./Neu./Neg.)& Avg. text length \\
 \hline
Amazon& 983  & 0.8993/0.0264/0.0743 &  49.6185 \\
SST& 1101  &  0.4033/0.2080/0.3887 & 19.3224  \\
\hline
\end{tabular}
\label{tb1}
\end{table}
In this paper, only a small number of samples selected from the original datasets were taken as the test sets. As shown in Table \ref{tb1}, 983 samples from the Amazon.com review dataset and 1101 samples from the SST dataset were used for testing. Looking at the distribution of positive/neutral/negative reviews, we see that the SST dataset has a more balanced distribution than the Amazon.com one, which contains a larger number of positive reviews and a fewer number of negative and neutral reviews. Moreover, Table \ref{tb1} also shows the average text length of these two testing sets. Amazon.com review data has a longer average text length, and the SST dataset usually has short reviews. To further discuss the two datasets, the statistics of their review text lengths are plotted in Fig. \ref{f7}.

From Fig. \ref{f7}, it is more clear that these two datasets have different distributions on review text length. Combined with the results in Table \ref{tb1}, we can say these two datasets consider both data balance and text length for testing the given AI product, which would be helpful for the complete evaluation in AIQM. 

\subsection{Evaluation metrics}
We evaluated the robustness of the AI  model by using the definition of an adversarial example in which a small perturbation in the data fools the model. Since the evaluation of using ChatGPT for sentiment analysis is essentially classification problem, we can apply two traditional classification metrics: accuracy ($Acc$) and attack success rate (ASR).
\begin{equation}
Acc=\frac{\text{\# of correctly classified samples}}{\text{\# of total testing samples}}\times 100\%
\end{equation}
\begin{equation}
ASR=\frac{\text{\# of successfully attaked samples}}{\text{\# of total testing samples}}\times 100\%
\end{equation}

Although these two metrics are similar, they have different meanings. $Acc$ gives the percentage of samples for which the decision matched the ground truth, and $ASR$ gives the difference in accuracy between before and after perturbation. Therefore, $ASR$ can be directly used for robustness analysis, meanwhile we can also compare accuracy before and after perturbation for robustness analysis. 
\subsection{Perturbation and robustness analysis} 
\begin{figure*}
\centering
\includegraphics[scale=0.5]{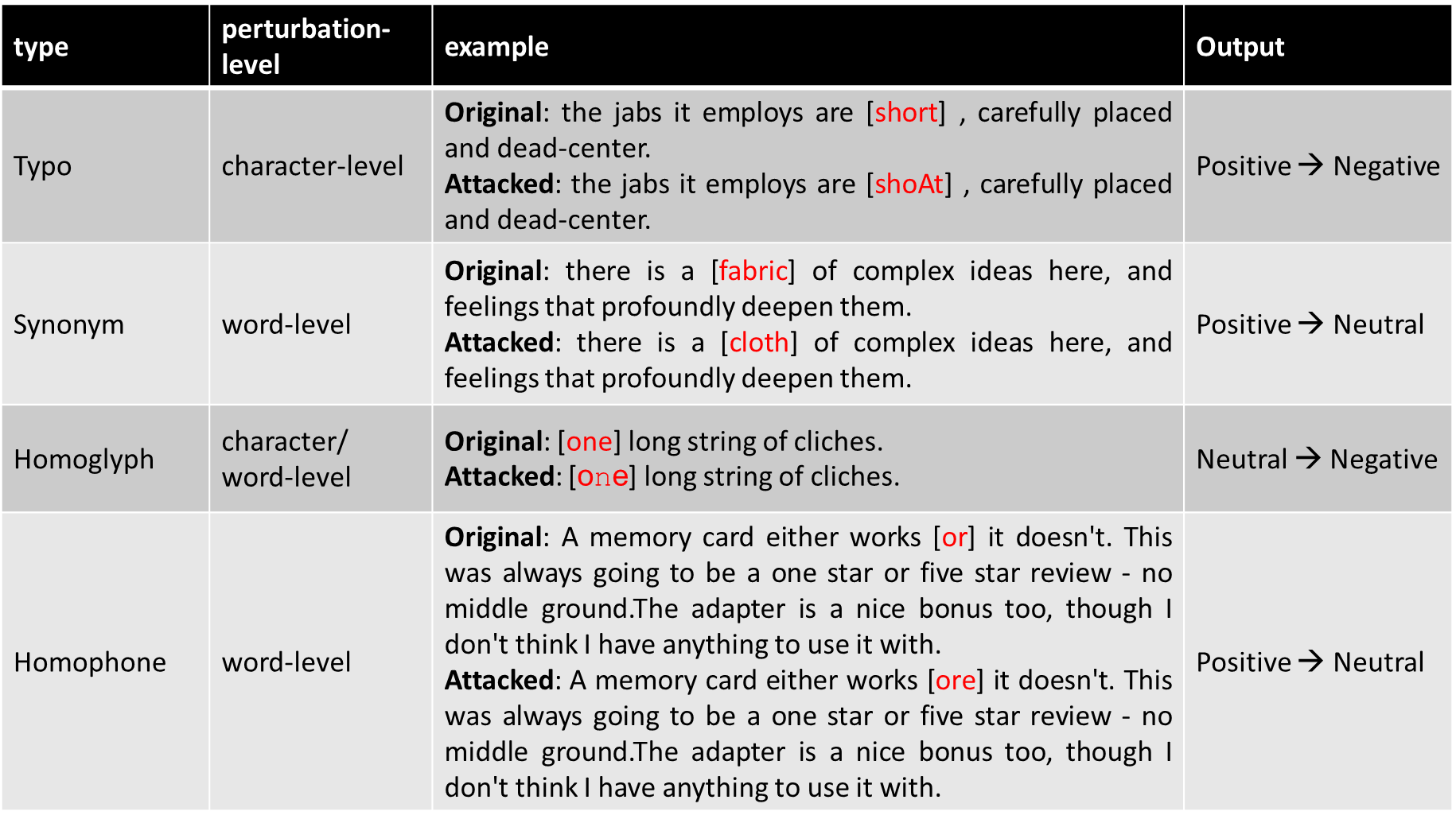}
\caption{The homoglyph of alphabets}
\label{f9}
\end{figure*}

As shown in Fig. \ref{f9}, this paper considered four types of perturbation from different perspectives, such as typo, synonym, homoglyph, and homophone perturbation. They mainly involve character- and word-level perturbation. To maintain naturalness and readability, the adversarial texts were generated in accordance with the given perturbation and kept within an edit distance of one word, as seen in the examples in Fig. \ref{f9}, where successfully attacked review comments via the given four perturbations are presented.

(1) typo perturbation

Psychological studies have shown that a sentence with one or more typos can often be still comprehended by a person \cite{b37}. However, in the information era with computers, words with typo perturbation are encoded differently, which may lead to incorrect machine processing. Therefore, typo perturbation is regarded as a common attack in NLP studies, mainly because typing errors are common in computer input.
In this paper, we used the four kinds of common typo perturbations for words introduced in the TextAttack package \cite{b38}: swapping two adjacent letters in a word, substituting a letter with a random letter, randomly deleting a letter, and randomly inserting a letter. Adversarial texts with these perturbations were generated by restricting the edit distance in the sentence to maintain semantic understanding, which we refer to as “one-word perturbation”. Typos were accordingly added to review comments in the datasets, and the performance of the AI model's robustness was tested, presented in Table \ref{tb2}.

\begin{table}
\centering
\caption{ROBUSTNESS AGAINST TYPO PERTURBATION}
\begin{tabular}{ccccc}
\hline
 & $ori_{acc}$ & $pert_{acc}$ & $\Delta_{diff}$& $ASR$ \\
 \hline
Amazon& 0.8942  &  0.7636 & 0.1306 & 0.1273\\
SST& 0.8065  &  0.6129  & 0.1936 & 0.1935  \\
\hline
\end{tabular}
\label{tb2}
\end{table}

The resulting performance is summarized in Table \ref{tb2}, where $ori_{acc}$ and $pert_{acc}$ represent accuracy before and after perturbation, respectively, and their difference is expressed as $\Delta_{diff}$. These results show that the ChatGPT-based sentiment analysis system performs relatively better on Amazon.com review data than on SST data. Moreover, the ASRs for the two datasets were close to the drops in accuracy. The larger $ASR$ for the SST dataset, along with the text length analysis results shown in Fig. \ref{f7}, indicates that short review texts are more easily attacked by typo perturbation. This is consistent with the common understanding that longer texts are more robust against attacks. 

(2) synonym perturbation

Another common attack used in NLP studies is synonym perturbation in which a word is replaced with a synonym. The text with the replacement word should have readability and meaning similar to those of the original text. However, in ML/DL-based NLP studies, an NLP model may behave differently on synonyms because the different words have different encodings in the tokenization.As a result, synonym perturbation has generally worked well in conventional adversarial text generation studies, especially in sentiment analysis where sentiment words play a crucial role. This paper adopts the provided synonym perturbation algorithm in \cite{b39} to create adversarial text in both the Amazon.com review dataset and the SST dataset. Results of robustness analysis are presented in Table \ref{tb3}.
\begin{table}
\centering
\caption{ROBUSTNESS AGAINST SYNONYM PERTURBATION }
\begin{tabular}{ccccc}
\hline
& $ori_{acc}$ & $pert_{acc}$ & $\Delta_{diff}$& $ASR$ \\
 \hline
Amazon& 0.8942   &  0.5781& 0.3161 &  0.3642
 \\
SST& 0.8065  &  0.3871  &0.4194 & 0.5200 \\
\hline
\end{tabular}
\label{tb3}
\end{table}

As shown in Table \ref{tb3}, the ASR values for synonym perturbation were much larger than those for typo perturbation, indicating that synonym perturbation is a stronger attack than a typo one. Again, the longer texts in the Amazon.com review dataset were more robust than the shorter ones in the SST dataset against synonym perturbation.

(3) homoglyph perturbation

\begin{figure}
\centering
\includegraphics[scale=0.6]{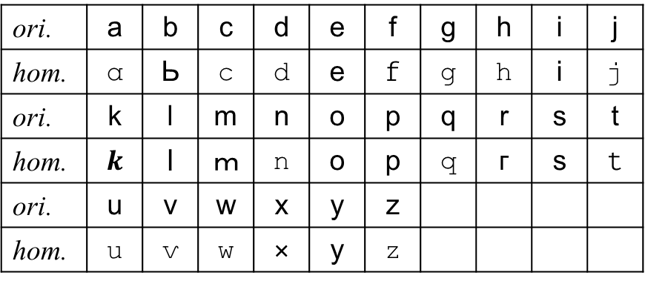}
\caption{Example homoglyphs (hom.) of 26 letters in English alphabet}
\label{f8}
\end{figure}

In \cite{b38}, Gao et al. also proposed using homoglyph perturbation for adversarial text attacks. The idea is to replace a character with a similar-looking character, e.g., using a symbol with an identical shape but with a different ASCII code, as illustrated by the examples in Fig. \ref{f8}. This perturbation was demonstrated to be useful in adversarial text attacks. In accordance with the number of replaced characters, homoglyph perturbation can be categorized as character-level (transforming a few characters) or word-level (transforming all characters in a word). Since homoglyph perturbation retains a look similar to that of the original word, it does not affect readability or the semantic meaning of the original text, which is useful in adversarial robustness studies. Therefore, in this paper, word-level homoglyph perturbation is adopted first. On the basis of word importance in sentiment analysis as determined using the Vader analyzer \cite{b40}, adversarial texts were generated to test the performance of ChatGPT on sentiment analysis. Table \ref{tb4} summarizes performance on the two datasets.

\begin{table}
\centering
\caption{PROBUSTNESS AGAINST HOMOGLYPH PERTURBATION}
\begin{tabular}{ccccc}
\hline
& $ori_{acc}$ & $pert_{acc}$ & $\Delta_{diff}$& $ASR$\\
 \hline
Amazon& 0.8942	&0.6536 & 0.2406 &0.2397
 \\
SST& 0.8065	&0.7419& 0.0646 &	0.1290
 \\
\hline
\end{tabular}
\label{tb4}
\end{table}

A comparison of the results in Table \ref{tb2} -\ref{tb4} shows that homoglyph perturbation is a stronger attack than typo perturbation but weaker than synonym perturbation. ChatGPT thus achieved relatively good robustness against homoglyph perturbation when used for sentiment analysis.

(4) homophone perturbation

In contrast to homoglyph perturbation in which text appearance is transformed, homophone perturbation \cite{b41} transforms text on the basis of pronunciation. Homophones, which are words with similar sounds but are spelled differently and have different meanings, can cause the system to misclassify the sentiment. It has been demonstrated to be useful, especially in adversarial Chinese text generation \cite{b41}. In this paper, we use this perturbation in English text perturbation. First, word importance for sentiment analysis was determined using the Vader analyzer. Then, homophone perturbation was used to generate adversarial texts based on the order of word importance. Robustness analysis based on these adversarial texts is done on two datasets, and results are presented in Table \ref{tb5}.
 
\begin{table}
\centering
\caption{PROBUSTNESS AGAINST homophone perturbation}
\begin{tabular}{ccccc}
\hline
& $ori_{acc}$ & $pert_{acc}$ & $\Delta_{diff}$& $ASR$\\
 \hline
Amazon& 0.8942	&0.8445 & 0.0497 & 0.0497 \\
SST& 0.8065	&0.7419& 0.0646 &	0.0645 \\
\hline
\end{tabular}
\label{tb5}
\end{table}
As shown in Table \ref{tb5}, the accuracy drop and ASR values for homophone perturbation were quite low, indicating that homophone perturbation is not a strong attack against a ChatGPT-based system and that ChatGPT is robust against homophone perturbation when used for sentiment analysis. 

(5) discussion

Looking at the results in Tables \ref{tb2}–\ref{tb5}, we see that synonym perturbation was the strongest type of attack as it resulted in the largest reduction in accuracy and largest ASR values on adversarial texts. A comprehensive evaluation shows that it is difficult to generate a strong attack against a ChatGPT-based system by using conventional methods, including character- and word-level. Therefore, we can say that the ChatGPT-based sentiment analysis system can be robust against adversarial text perturbations in real applications of sentiment analysis.   

\section{Conclusion}
This paper introduces a sentiment-analysis system based on ChatGPT as the studied product for AI quality assurance. Meanwhile, the study mainly focuses on the specific quality assurance of stability. Two topics are discussed to examine the stability issue. One is based on the operation uncertainty, and several factors are analyzed and discussed. The second one is based on ChatGPT's robustness against four types of perturbations. Two benchmark datasets were used for the evaluation. Results demonstrated that the ChatGPT-based sentiment analysis system is robust against all four perturbations although a bit weaker against synonym perturbation. Through the results, we can conclude that it is reasonable to get a robust sentiment analysis product based on ChatGPT. Still, it is also necessary to notice the operation uncertainty due to ChatGPT continuously updating itself, the effect of differences in the time of operation, and some other factors.

\section{Acknowledgement}
This research is supported by the project ’JPNP20006’, commissioned by the New Energy and Industrial Technology Development Organization (NEDO). 
and partly supported by JSPS Grant-in-Aid for Early-Career Scientists (Grant Number 22K17961)


\begin{thebibliography}{00}
\bibitem{b1} Ouyang, Tinghui, Yoshinao Isobe, Saima Sultana, Yoshiki Seo, and Yutaka Oiwa. "Autonomous driving quality assurance with data uncertainty analysis." In 2022 International Joint Conference on Neural Networks (IJCNN), pp. 1-7. IEEE, 2022.
\bibitem{b2} Shinde, Pramila P., and Seema Shah. "A review of machine learning and deep learning applications." In 2018 Fourth international conference on computing communication control and automation (ICCUBEA), pp. 1-6. IEEE, 2018.
\bibitem{b3} Hordri, Nur Farhana, Siti Sophiayati Yuhaniz, and Siti Mariyam Shamsuddin. "Deep learning and its applications: a review." In Conference on Postgraduate Annual Research on Informatics Seminar, pp. 1-5. 2016.
\bibitem{b4} Machine Learning Quality Management Guideline$(https://www.digiarc.aist.go.jp/en/publication/aiqm/)$
\bibitem{b5} Wang, Jiaqi, Zhengliang Liu, Lin Zhao, Zihao Wu, Chong Ma, Sigang Yu, Haixing Dai et al. "Review of large vision models and visual prompt engineering." arXiv preprint arXiv:2307.00855, 2023.
\bibitem{b6} Rishi Bommasani, Drew A Hudson, Ehsan Adeli, Russ Altman, Simran Arora, Sydney von Arx, Michael S Bernstein, Jeannette Bohg, Antoine Bosselut, Emma Brunskill, et al.  On the opportunities and risks of foundation models. arXiv preprint arXiv:2108.07258, 2021.
\bibitem{b7} OpenAI. $https://chat.openai.com.chat$, 2023.
\bibitem{b8} Long Ouyang, Jeff Wu, Xu Jiang, Diogo Almeida, Carroll L Wainwright,
Pamela Mishkin, Chong Zhang, Sandhini Agarwal, Katarina Slama, Alex
Ray, et al. Training language models to follow instructions with human
feedback. arXiv preprint arXiv:2203.02155, 2022.
\bibitem{b9} Jiao, Wenxiang, Wenxuan Wang, Jen-tse Huang, Xing Wang, and Zhaopeng Tu. "Is ChatGPT a good translator? A preliminary study." arXiv preprint arXiv:2301.08745, 2023.
\bibitem{b10} Wu, Haoran, Wenxuan Wang, Yuxuan Wan, Wenxiang Jiao, and Michael Lyu. "Chatgpt or grammarly? evaluating chatgpt on grammatical error correction benchmark." arXiv preprint arXiv:2303.13648, 2023.
\bibitem{b11} Bordt, Sebastian, and Ulrike von Luxburg. "Chatgpt participates in a computer science exam." arXiv preprint arXiv:2303.09461 , 2023.
\bibitem{b12} Selivanov, Alexander, Oleg Y. Rogov, Daniil Chesakov, Artem Shelmanov, Irina Fedulova, and Dmitry V. Dylov. "Medical image captioning via generative pretrained transformers." Scientific Reports 13, no. 1 : 4171, 2023.
\bibitem{b13} Mitrović, Sandra, Davide Andreoletti, and Omran Ayoub. "Chatgpt or human? detect and explain. explaining decisions of machine learning model for detecting short chatgpt-generated text." arXiv preprint arXiv:2301.13852, 2023.
\bibitem{b14} Dai, Haixing, Zhengliang Liu, Wenxiong Liao, Xiaoke Huang, Zihao Wu, Lin Zhao, Wei Liu et al. "Chataug: Leveraging chatgpt for text data augmentation." arXiv preprint arXiv:2302.13007, 2023.
\bibitem{b15} Frieder, Simon, Luca Pinchetti, Ryan-Rhys Griffiths, Tommaso Salvatori, Thomas Lukasiewicz, Philipp Christian Petersen, Alexis Chevalier, and Julius Berner. "Mathematical capabilities of ChatGPT." arXiv preprint arXiv:2301.13867, 2023.
\bibitem{b16} Liu, Hanmeng, Ruoxi Ning, Zhiyang Teng, Jian Liu, Qiji Zhou, and Yue Zhang. "Evaluating the logical reasoning ability of chatgpt and gpt-4." arXiv preprint arXiv:2304.03439, 2023.
\bibitem{b17} Wang, Jindong, Xixu Hu, Wenxin Hou, Hao Chen, Runkai Zheng, Yidong Wang, Linyi Yang et al. "On the robustness of chatgpt: An adversarial and out-of-distribution perspective." arXiv preprint arXiv:2302.12095, 2023.
\bibitem{b18} Wang, Boxin, Weixin Chen, Hengzhi Pei, Chulin Xie, Mintong Kang, Chenhui Zhang, Chejian Xu et al. "DecodingTrust: A Comprehensive Assessment of Trustworthiness in GPT Models." arXiv preprint arXiv:2306.11698 , 2023.
\bibitem{b19} Jones, Erik, Anca Dragan, Aditi Raghunathan, and Jacob Steinhardt. "Automatically Auditing Large Language Models via Discrete Optimization." arXiv preprint arXiv:2303.04381, 2023.
\bibitem{b20} Xu, Xilie, Keyi Kong, Ning Liu, Lizhen Cui, Di Wang, Jingfeng Zhang, and Mohan Kankanhalli. "An LLM can Fool Itself: A Prompt-Based Adversarial Attack." arXiv preprint arXiv:2310.13345, 2023.
\bibitem{b21} Reiss, Michael V. "Testing the reliability of chatgpt for text annotation and classification: A cautionary remark." arXiv preprint arXiv:2304.11085, 2023.
\bibitem{b22} Rozado, David. "The political biases of chatgpt." Social Sciences 12, no. 3 : 148, 2023.
\bibitem{b23} Li, Tsz-On, Wenxi Zong, Yibo Wang, Haoye Tian, Ying Wang, and Shing-Chi Cheung. "Finding Failure-Inducing Test Cases with ChatGPT." arXiv preprint arXiv:2304.11686, 2023.
\bibitem{b24}Borji, Ali. "A categorical archive of ChatGPT failures." arXiv preprint arXiv:2302.03494, 2023.
\bibitem{b25}Yang X, Li Y, Zhang X, Chen H, Cheng W. Exploring the limits of chatgpt for query or aspect-based text summarization. arXiv preprint arXiv:2302.08081, 2023 Feb 16.
\bibitem{b26} Jalil, Sajed, Suzzana Rafi, Thomas D. LaToza, Kevin Moran, and Wing Lam. "Chatgpt and software testing education: Promises and perils." arXiv preprint arXiv:2302.03287, 2023.
\bibitem{b27} Wang, Boxin, Chejian Xu, Shuohang Wang, Zhe Gan, Yu Cheng, Jianfeng Gao, Ahmed Hassan Awadallah, and Bo Li. "Adversarial glue: A multi-task benchmark for robustness evaluation of language models." arXiv preprint arXiv:2111.02840, 2021.
\bibitem{b28} Ouyang, Tinghui, Yoshiki Seo, and Yutaka Oiwa. "Quality assurance study with mismatched data in sentiment analysis." In 2022 29th Asia-Pacific Software Engineering Conference (APSEC), pp. 442-446. IEEE, 2022.
\bibitem{b29} Zhang, Wenxuan, Yue Deng, Bing Liu, Sinno Jialin Pan, and Lidong Bing. "Sentiment Analysis in the Era of Large Language Models: A Reality Check." arXiv preprint arXiv:2305.15005, 2023.

\bibitem{b30} Puigcerver, Joan, Carlos Riquelme, Basil Mustafa, and Neil Houlsby. "From sparse to soft mixtures of experts." arXiv preprint arXiv:2308.00951, 2023.
\bibitem{b31} $https://community.openai.com/t/chatgpt-results-much-better-than-api/336749$
\bibitem{b32} $https://community.openai.com/t/different-output-generated-for-same-prompt-in-chat-mode-and-api-mode-using-gpt-3-5-turbo/318246$
\bibitem{b33} Chen, Lingjiao, Matei Zaharia, and James Zou. "How is ChatGPT's behavior changing over time?." arXiv preprint arXiv:2307.09009, 2023.
\bibitem{b34} IEEE standard glossary of software engineering terminology. IEEE Std 610.12-1990, pages 1–84, Dec 1990.
\bibitem{b35} McAuley, Julian, and Jure Leskovec. "Hidden factors and hidden topics: understanding rating dimensions with review text." In Proceedings of the 7th ACM conference on Recommender systems, pp. 165-172. 2013.
\bibitem{b36} Pang, Bo, and Lillian Lee. "Seeing stars: Exploiting class relationships for sentiment categorization with respect to rating scales." arXiv preprint cs/0506075 , 2005.
\bibitem{b37} Rayner, Keith, Sarah J. White, and S. P. Liversedge. "Raeding wrods with jubmled lettres: There is a cost.", 2006.
\bibitem{b38} Gao, Ji, Jack Lanchantin, Mary Lou Soffa, and Yanjun Qi. "Black-box generation of adversarial text sequences to evade deep learning classifiers." In 2018 IEEE Security and Privacy Workshops (SPW), pp. 50-56. IEEE, 2018.
\bibitem{b39} Ren, Shuhuai, Yihe Deng, Kun He, and Wanxiang Che. "Generating natural language adversarial examples through probability weighted word saliency." In Proceedings of the 57th annual meeting of the association for computational linguistics, pp. 1085-1097. 2019.
\bibitem{b40} Hutto, Clayton, and Eric Gilbert. "Vader: A parsimonious rule-based model for sentiment analysis of social media text." In Proceedings of the international AAAI conference on web and social media, vol. 8, no. 1, pp. 216-225. 2014.
\bibitem{b41} Xu, En-Hui, Xiao-Lin Zhang, Yong-Ping Wang, Shuai Zhang, Li-Xin Liu, and Li Xu. "Adversarial Examples Generation Method for Chinese Text Classification." International Journal of Network Security 24, no. 4 (2022): 587-596.

\end{thebibliography}
\end{document}